\pdfoutput=1
\documentclass[11pt]{article}
\usepackage[table,xcdraw]{xcolor}

\usepackage[preprint]{acl}

\usepackage{times}
\usepackage{latexsym}

\usepackage{caption}
\usepackage{graphicx}
\usepackage{float} 
\usepackage{subcaption}

\usepackage{amsmath}
\usepackage{amsfonts}

\usepackage{multirow}
\usepackage[T1]{fontenc}
\usepackage[utf8]{inputenc}

\usepackage{microtype}

\usepackage{inconsolata}

\usepackage{graphicx}

\title{Unveiling the Impact of Multimodal Features on Chinese Spelling Correction: From Analysis to Design}

\author{
  Xiaowu Zhang$^1$ \quad Hongfei Zhao$^2$ \quad Jingyi Hou$^1$ \quad Zhijie Liu$^1$ \\
  $^1$University of Science and Technology Beijing \\
  $^2$Fudan University, Shanghai 200433, China \\
  \texttt{zhangxw21@outlook.com} \quad \texttt{iioSnail@163.com} \\
  \texttt{houjingyi@ustb.edu.cn} \quad \texttt{liuzhijie2012@gmail.com}
}

\begin{document}
\maketitle
\begin{abstract}
The Chinese Spelling Correction (CSC) task focuses on detecting and correcting spelling errors in sentences. Current research primarily explores two approaches: traditional multimodal pre-trained models and large language models (LLMs). However, LLMs face limitations in CSC, particularly over-correction, making them suboptimal for this task. While existing studies have investigated the use of phonetic and graphemic information in multimodal CSC models, effectively leveraging these features to enhance correction performance remains a challenge. To address this, we propose the Multimodal Analysis for Character Usage (\textbf{MACU}) experiment, identifying potential improvements for multimodal correctison. Based on empirical findings, we introduce \textbf{NamBert}, a novel multimodal model for Chinese spelling correction. Experiments on benchmark datasets demonstrate NamBert's superiority over SOTA methods. We also conduct a comprehensive comparison between NamBert and LLMs, systematically evaluating their strengths and limitations in CSC. Our code and model are available at https://github.com/iioSnail/NamBert.
\end{abstract}

\section{Introduction}

\begin{figure}[t]
  \centering
  \includegraphics[width=1\linewidth, trim=0 200 0 50, clip]{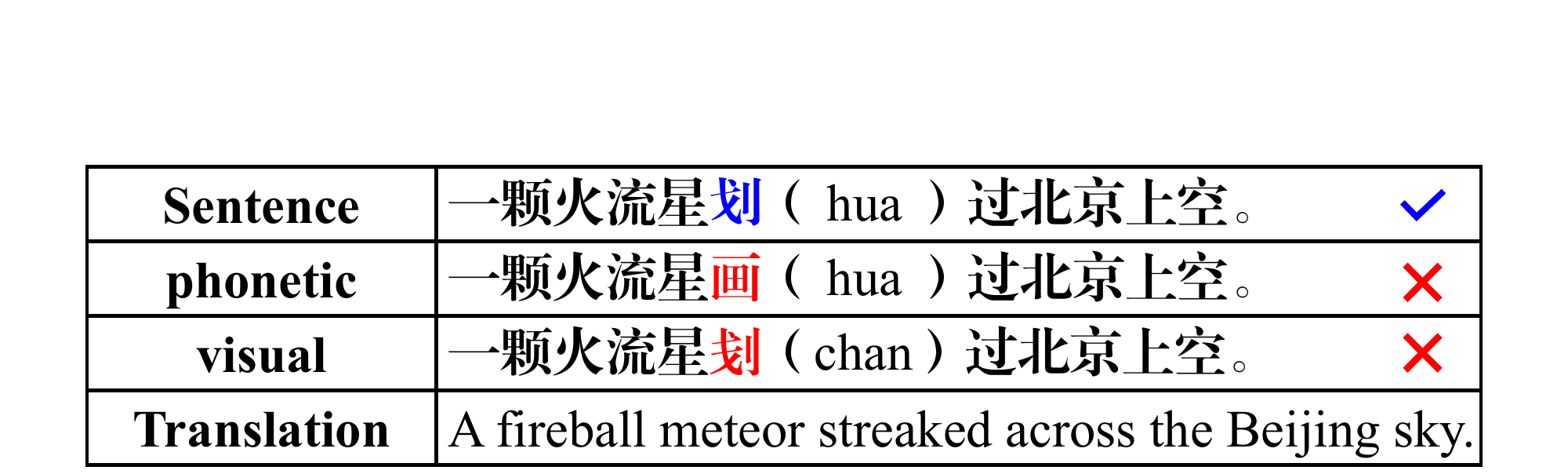}
  \caption[]{Examples of Chinese spelling errors. Mis-spelling characters are marked in \textcolor{red}{red}, while the correct characters are marked in \textcolor{blue}{blue}, with the corresponding phonics provided in brackets.}
  \label{fig0}
\end{figure}

The primary objective of Chinese Spelling Correction is to detect erroneous characters in sentences and provide the correct corrections. As a crucial task in the field of Natural Language Processing (NLP) \cite{p1n0}, CSC plays a key role in various NLP applications \cite{p1n1,p1n2,p1n3}. Chinese spelling errors are typically caused by the misuse of homophones (characters with similar pronunciations) and visually similar characters \cite{p1n4,p1n5}. Figure \ref{fig0} illustrates the two most common types of errors in the CSC task.

%Fig1

In recent years, the emergence of large language models has introduced new solutions for the CSC task. However, studies have shown that LLMs suffer from slow inference speed and over-correction issues \cite{a002}, leading to unstable correction performance overall \cite{a000}. In contrast, CSC models that incorporate multimodal information have demonstrated more stable performance \cite{a001}. Research indicates that integrating phonetic and graphemic information can significantly enhance the performance of Chinese spelling correction models \cite{p1n6}. Consequently, mainstream CSC approaches adopt various strategies to fuse these two modalities to improve correction accuracy \cite{p1n7}. For instance, \citet{p1n8} employs a four-layer Transformer encoder and a four-layer convolutional neural network to extract phonetic and graphemic features, which are then combined with Bert for semantic modeling \cite{p1n85}. \citet{p1n9} encodes Pinyin and graphemic information using a GRU network and integrates multimodal features at the word embedding layer before feeding them into Bert for further feature extraction. \citet{p1n10} utilizes an encoder along with two parallel decoders, one for predicting target characters and the other for their corresponding phonetic information to enhance correction accuracy.

Although different models adopt various fusion strategies for phonetic and graphemic information, existing research consistently demonstrates that incorporating multimodal information enhances the correction performance of CSC models. To investigate whether pre-trained models genuinely encode phonetic and graphemic features, \citet{p3n1} proposes a Probe Task to analyze the encoding of phonetic and graphemic information in pre-trained models and designed the CCCR task to evaluate how models utilize erroneous character information during the correction process. However, there are significant differences in how different CSC approaches utilize phonetic and graphemic information in practical applications. Therefore, efficient utilization of multimodal information remains a key issue. In response to this, this paper discusses the following two questions:

\textbf{Multimodal models with different structures?} We designed the Multimodal Analysis for Character Usage task (MACU) exploration experiment to thoroughly analyze the characteristics of different Chinese spelling correction models and their ability to utilize phonetic and graphemic information.

\textbf{How can phonetic and graphemic information be effectively modeled to enhance the performance of multimodal spelling correction?} We conducted a series of experiments based on ChineseBERT, exploring methods to optimize the model structure and improve its correction performance.

Based on the results of the exploration experiment, we propose the following optimization strategies: (1) Use a non-aligned multimodal fusion method to reduce the loss of multimodal information. (2) Use a post-fusion approach to integrate multimodal information, ensuring the prediction layer obtains more information about incorrect characters. (3) Optimize the loss function to enhance the model's focus on incorrect characters. %(4) Use a more efficient phonetic and graphemic encoding network that reduces computational costs while maintaining strong representation capabilities. At the same time, optimize the semantic encoder to improve further the model's ability to apply incorrect character information.

Furthermore, we conducted comparison experiments with LLMs using prompting strategies for error correction, analyzing the advantages and disadvantages of LLMs compared to traditional multimodal error correction solutions. Through this comparative analysis, we hope to provide new insights into Chinese spelling correction and contribute to its research and development.

\section{Related Work}
Early Chinese spelling correction tasks primarily utilized rule-based methods, relying on predefined linguistic rules or common spelling error cases for correction. However, their limited domain generalization and narrow error coverage led to significantly constrained correction capabilities. \citet{p2n1,p2n2} addresses various types of spelling errors by designing different rules and employing N-gram language models. \citet{p2n3} treats the Chinese spelling correction task as a sequence labeling task. Copy mechanisms have also been used in sequence-to-sequence frameworks, with the core idea of copying candidate correction words from a confusion set \cite{p2n4}.

%fig1.5
\begin{figure*}[ht]
  \centering
  \includegraphics[width=1\textwidth, trim=50 90 50 80, clip]{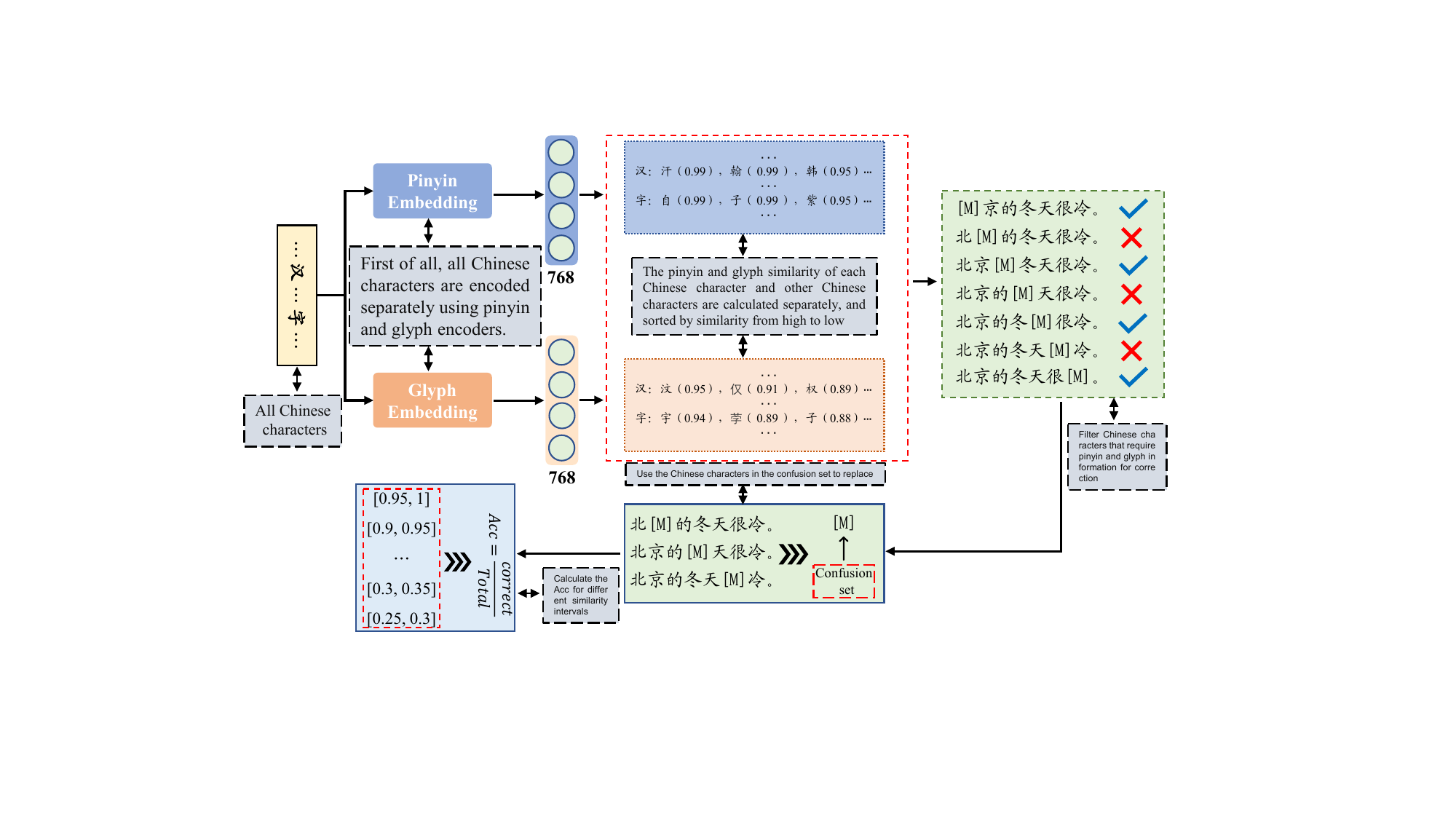}
  \caption[]{The encoder extracts Chinese characters' phonetic and graphical features separately and constructs a confusion set. Then, characters in the Chinese text are selected and replaced, and the model’s prediction accuracy for these characters is calculated. The figure shows the character replacement process based on phonetic and graphical similarity to test the model’s performance within different similarity ranges.}
  \label{fig15}
\end{figure*}

With technological advancements, rule-based and statistical methods were gradually abandoned due to their complexity and high correction costs \cite{p2n10,p2n12}. Deep learning-based methods gradually became the primary solution for Chinese spelling correction \cite{p2n13,p2n14}. \citet{p2n5} identifies deficiencies in the error detection capabilities of pre-trained models and proposed an architecture comprising an error detection network and a correction network, where soft-mask detection results are fed into a Bert-based correction network. \citet{p2n6} incorporates phonetic information into word embeddings and employed dynamic programming algorithms along with phonetic similarity to address the issue of incoherent word predictions in previous models. \citet{p2n7} introduces an uncertainty-guided multimodal feature fusion strategy, dynamically integrating phonetic and graphemic information to effectively enhance spelling correction performance, significantly outperforming previous multimodal models.\citet{p2n8} proposes a framework called uChecker for unsupervised spelling error detection and correction, introducing a confusion set strategy to fine-tune masked language models, thereby enhancing unsupervised correction performance. \citet{p2n9} proposes a novel knowledge graph-based correction method, injecting queried triples as domain knowledge into sentences, enabling the model to possess reasoning abilities and common sense. \citet{p2n11} addresses issues related to Chinese spelling correction datasets and performance bottlenecks of current models, proposing relevant solutions.

\section{Multimodal Analysis for Character Usage}

%fig3
\begin{figure*}[ht]
\centering
\subfloat[Phonetic result.]{\includegraphics[width=3in]{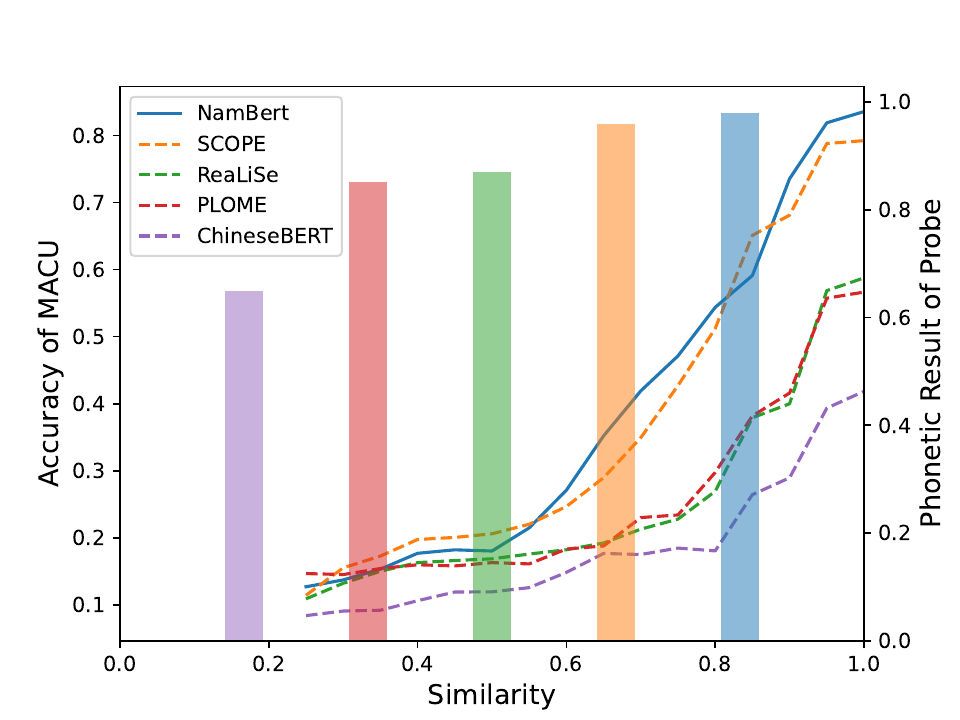}%
\label{fig2a}}
\hfil
\subfloat[Graphemic result.]{\includegraphics[width=3in]{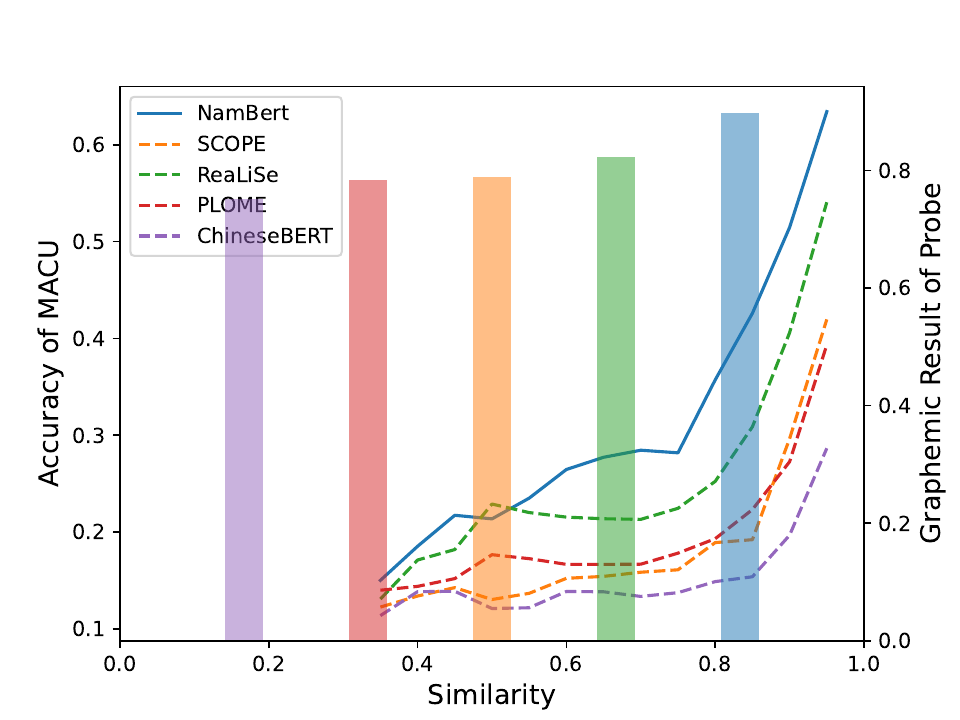}%
\label{fig2b}}
\caption{The figure shows the results of the probe experiment and MACU experiment. The bar chart represents the results of the probe experiment, where "Similarity" denotes the lower bound of the similarity range. The line chart shows the accuracy of the MACU experiment at that similarity level.}
\label{fig2}
\end{figure*}

In recent years, various multimodal models for Chinese spelling correction have emerged. However, the actual correction accuracy of these models has not seen substantial improvement. Overcoming the bottleneck of multimodal correction models and exploring potential directions for improvement have become urgent issues to address. This paper designs the MACU task to analyze the phonetic-graphemic information utilization ability of multimodal spelling correction models, aiming to investigate the strengths and weaknesses of different multimodal Chinese spelling correction models.

Specifically, we first encode the phonetic and graphemic information of all Chinese characters, resulting in 768-dimensional phonetic and graphemic embeddings. We use the cosine similarity metric to evaluate the phonetic and graphemic similarity between characters. Then, we apply normalization to map the similarity values to the [0, 1] range.

%eq1
\begin{equation}
C_{sim}  = \left ( \frac{h_{x}\cdot h_{y}  }{\left \| h_{x} \right \| \left \| h_{y} \right \|  }  \right )
\label{eq1}
\end{equation}

\noindent where $C_{sim}$ represents the final similarity result, and $h_{x}$ and $h_{y}$ represent the two features for which the similarity is being calculated.

Next, we sort the characters in the confusion set by their similarity probability, from high to low, resulting in two confusion sets with similarity ranges between [0, 1]: the only phonetic-similar confusion set ($C_{p}$) and the only graphemic-similar confusion set ($C_{g}$). To ensure that the differences in the correction capabilities of the experimental models do not affect the results, we selected 3,000 samples from the training set that all the models commonly used and apply a diagonal masking pattern to mask each character in the sentence one by one, using the MacBert \cite{a003} for preliminary predictions. Through this process, we filter out the characters that require additional phonetic and graphemic information to correct and construct a test set $S$ containing 14,937 test samples. To evaluate the model's performance across different difficulty levels, we divide $C_{p}$ and $C_{g}$ into 20 intervals, each with a 0.05 range, and randomly select character pairs within each interval to replace masked characters in set $S$. This allows us to test the model's performance within different similarity ranges. Finally, we calculate the accuracy metric for each model within different ranges using Equation \ref{eq2}.

%eq2
\begin{equation}
Acc_r= \frac{C_r}{T_r}
 \label{eq2}
\end{equation}

\noindent Here, $T_r$ refers to the total number of substituted characters, and $C_r$ indicates the number of characters correctly corrected by the model in the range of $r$.

This paper conducts exploratory experiments on the multimodal correction models ReaLiSe \cite{p1n8}, SCOPE \cite{p1n10}, PLOME \cite{p1n9}, and ChineseBERT \cite{p46n1} and systematically studies and analyzes the relationship between the encoding ability and utilization ability of multimodal information, in combination with the probing experiment method proposed by \citet{p3n1}.

%tab1
\begin{table}[]
\centering
\resizebox{0.8\linewidth}{!}{ 
\begin{tabular}{ccc}
\hline
\textbf{Model}      & \textbf{\textbf{}$P_{probe}$} & \textbf{$P_{MACU}$} \\ \hline
ChineseBERT &65.0    & 21.9   \\ 
PLOME      &85.2    & 30.4  \\ 
ReaLiSe     &87.1    & 30.2   \\ 
SCOPE       &95.9    & 45.5   \\ \hline
NamBert     &98.0    & 47.4   \\ \hline
\end{tabular}
}
\caption{The results of the model's probe experiment and MACU  experiment are shown. $P_{probe}$ refers to the phonetic results of the probe experiment, and $P_{MACU}$ represents the weighted avarage accuracy of MACU.}
\label{tb1}
\end{table}

Based on the model structure and the experimental results from Figure \ref{fig2}, the following conclusions can be drawn: (1) multimodal information is crucial for the CSC task. The stronger the encoding ability of the phonetic and graphemic shape information, the higher the model's correction accuracy. Further analysis reveals that as the model's ability to encode phonetic and graphemic information improves, the performance of the multimodal correction model in the MACU task is also optimized accordingly. This indicates that obtaining richer multimodal features during the prediction phase allows for more accurate correction of spelling errors that rely on phonetic and graphemic information. (2) The method of multimodal fusion has an impact on the model's correction ability. Compared to ReaLiSe, PLOME adopts an early fusion approach for multimodal information. While the model learns phonetic and graphemic information to some extent, it reduces the utilization of phonetic and graphemic information in the prediction layer. This suggests that a late fusion method is more effective for CSC tasks.

%tab1
\begin{table}[]
\centering
\resizebox{0.8\linewidth}{!}{ 
\begin{tabular}{ccc}
\hline
\textbf{Model}      & \textbf{$G_{probe}$} & \textbf{$G_{MACU}$} \\ \hline
ChineseBERT &75.1    & 16.0   \\ 
PLOME      &78.3    & 20.2  \\ 
ReaLiSe     &82.3    & 27.9   \\ 
SCOPE       &78.8    & 21.1   \\ \hline
NamBert     &89.8    & 34.6   \\ \hline
\end{tabular}
}
\caption{The results of the model's probe experiment and MACU  experiment are shown. $G_{probe}$ refers to the graphemic results of the probe experiment, and $G_{MACU}$ represents the weighted avarage accuracy of MACU.}
\label{tb12}
\end{table}

We introduce the model encoding ability metric, which is used to comprehensively evaluate the utilization ability of multimodal correction models for phonetic and graphemic information. By quantifying this encoding ability metric, we explore potential directions for improving the performance of multimodal Chinese spelling correction models. We divide the similarity range into n intervals with a step size of 0.05, where the lower boundary of each interval is denoted as $\phi_{i }$. Therefore, the weight for each point can be calculated using Equation \ref{eq3}:

%eq3
\begin{equation}
w_{i}=  \frac{\phi_{i } }{ {\textstyle \sum_{i\le n}^{} \phi_{i }}} 
 \label{eq3}
\end{equation}

We then use a weighted average to calculate the final MACU value.

%eq4
\begin{equation}
\tilde{A} _{MACU} = \textstyle \sum_{i\le n}^{} a_{i} \cdot w_{i}
\label{eq4}
\end{equation}

\noindent Here, $a_{i}$ represents the accuracy for interval $i$.

As shown in Table \ref{tb1} and Table \ref{tb12}, models with stronger modal information encoding ability also exhibit higher utilization of modal information. Therefore, in model design, how to effectively retain more modal information and pass it to the prediction layer becomes key to improving model performance. To further explore the performance of models under different structures, we conducted exploratory experiments on the ChineseBERT model to investigate potential improvement directions for multimodal correction models.

%tab2
\begin{table}[h]
\centering
\resizebox{0.8\linewidth}{!}{ 
\begin{tabular}{ccc}
\hline
\textbf{Model}   & \textbf{$P_{MACU}$} & \textbf{$G_{MACU}$} \\ \hline

w/ Posterior Fusion    & 23.7             & 19.1           \\ 
w/ Align          & 18.4             & 14.3            \\ \hline
ChineseBERT       & 21.9            & 16.0            \\ \hline
\end{tabular}
}
\caption{The table shows the performance of ChineseBERT after retraining with different modal structures on the MACU task. "Posterior Fusion" indicates changing the modal front fusion to posterior fusion, while "Align" refers to fusing modal features by addition.}
\label{tb2}
\end{table}

As shown in Table \ref{tb2}, based on the results of the exploratory models, we conducted a comprehensive analysis of the impact of different schemes on the ChineseBERT model. The experiments demonstrate that when the modality fusion method of ChineseBERT is changed to late fusion, the utilization of phonetic and graphemic information improves. The late fusion approach more effectively retains the multimodal information of incorrect characters, significantly enhancing error correction accuracy. However, when using non-aligned multimodal information fusion, the strong overlap between feature information leads to a feature coverage issue, causing information loss and negatively affecting the model's performance, resulting in suboptimal results. 

\section{Non-aligned Multimodal BERT}
\begin{figure*}[t]
  \centering
  \includegraphics[width=0.9\textwidth, trim=20 40 20 40, clip]{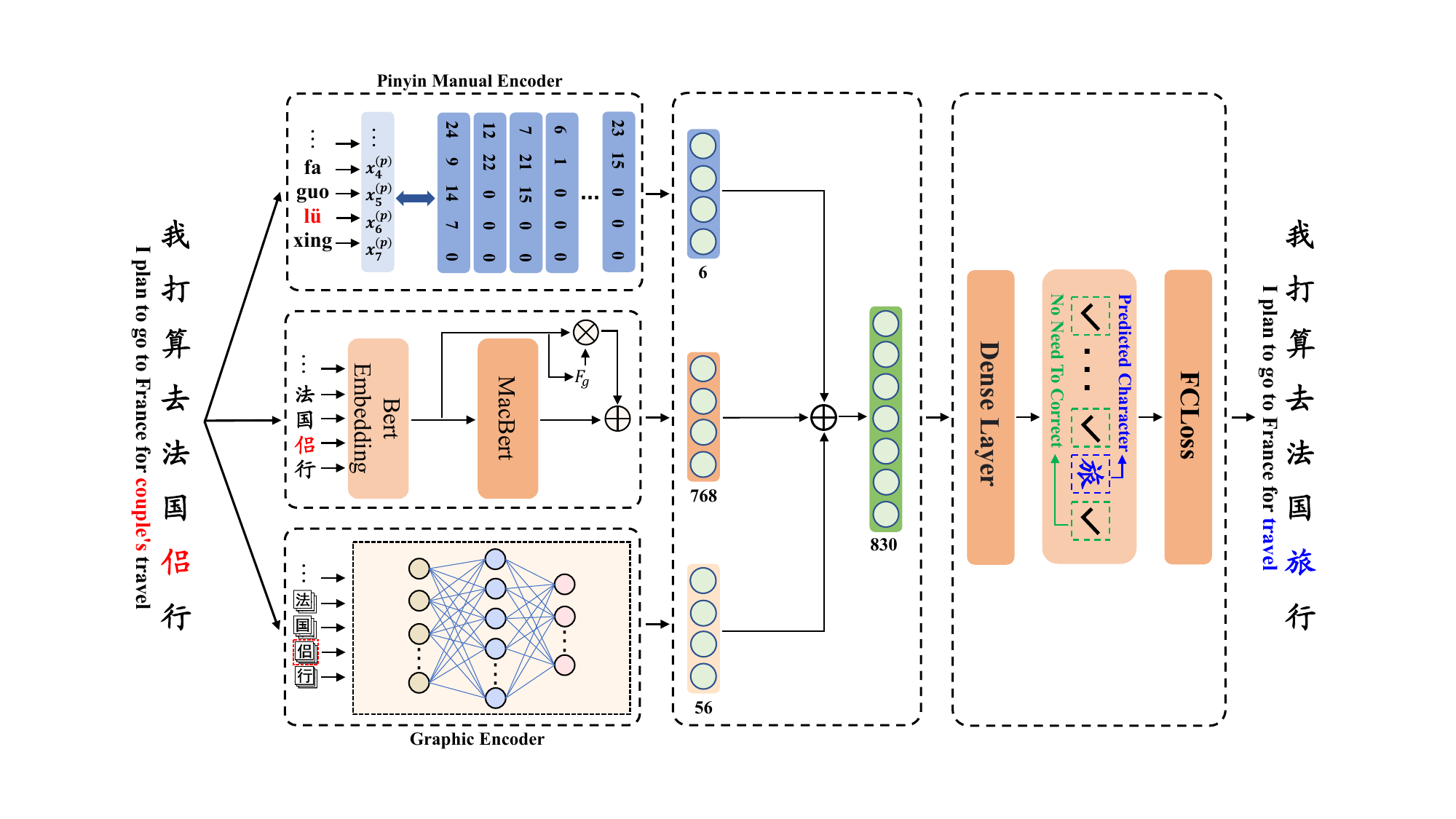}
  \caption[]{The architecture of NamBert. Multimodal information is extracted through a redesigned phonetic encoder, graphemic encoder, and semantic encoder. Modal information is used using a non-aligned posterior fusion approach, which is linearly transformed into 768 dimensions through a linear layer. The output layer fixes index 1 for the correct characters, while for incorrect characters, it outputs the corresponding index in the dictionary. Focal Loss is used to reduce the weight of index 1 so that the training focuses more on incorrect characters.}
  \label{fig3}
\end{figure*}

In this paper, based on the results of exploratory experiments, we designed the multimodal Chinese spelling correction model NamBert (\textbf{N}on-\textbf{a}ligned \textbf{m}ultimodal \textbf{BERT}), as shown in Figure \ref{fig3}. This model optimizes the encoder structure and introduces a non-aligned multimodal feature fusion mechanism, which maximally preserves the information from each modality through a late fusion mechanism. Additionally, We adopted a novel output method and introduced Focal Loss to the CSC task for the first time, enabling the model to more effectively utilize the multimodal information of incorrect characters, thereby significantly improving overall performance.

%fig3

% %fig4
% \begin{figure}[t]
%   \centering
%   \includegraphics[width=0.9\linewidth, trim=20 60 20 50, clip]{imgs/pic4.pdf}
%   \caption[]{A schematic diagram of phonetic encoding in
% NamBert.}
%   \label{fig4}
% \end{figure}

\textbf{Phonetic Encoder:} NamBert adopts a low-dimensional and efficient encoding method for phonetic features. Specifically, NamBert's phonetic encoder map each pinyin to a 6-dimensional vector. First, the PyPinyin \footnote{\url{https://github.com/mozillazg/python-pinyin}} converts Chinese characters into pinyin, and then a numeric mapping is assigned to each pinyin character. Next, each pinyin is transformed into a vector of a maximum length of 6 dimensions, and for pinyin with fewer than six characters, zeros are used for padding. The encoded phonetic information is then output through a linear transformation layer.

\textbf{Graphemic Encoder:} NamBert adopts three-layer feedforward neural network as the graphemic encoder. First, each Chinese character in the input sentence is converted into a 32×32 pixel image. These images are then processed in batches and fed into the graphemic encoder for feature extraction. To ensure that NamBert can fully learn the graphemic features, we also designed an innovative graphemic pretraining task.

\textbf{Semantic Encoder:} To preserve more of the original error character information during the prediction phase, the semantic encoder integrates the original word embeddings of the corresponding Chinese characters into the semantic features generated by Bert, thereby reducing the risk of misjudgment in the prediction layer. However, directly introducing word embedding features may result in excessive coverage of semantic features, causing unnecessary modifications to correct characters originally. To address this issue, we introduce a forget gate before feature fusion to dynamically select word embedding features and control the effective transfer of information. Specifically, given an input text sequence $X=\{x_0, x_1, \cdots, x_n\}$, word embeddings are first generated using Bert's Embedding layer, resulting in the embedding sequence $E=\{e_0, e_1, \cdots, e_n\}$.

%eq5
\begin{equation}
    \label{eq34}
    E = X W_e
\end{equation} 

The word embedding layer consists of an embedding matrix $W_e$ without bias. After obtaining embeddings $E$, they are fed into the Bert model to extract semantic features, resulting in the feature vector $H=\{h_0, h_1, \cdots, h_n\}$.

\begin{equation}
    \label{eq35}
    H = \textbf{Bert}(E)
\end{equation} 

The word embeddings are then filtered through a forget gate, resulting in the filtered embeddings $E'=\{e'_0, e'_1, \cdots, e'_n\}$.

\begin{equation}
    \label{eq36}
    \begin{aligned}
    E' & = F_g(E) \\
       & = \sigma(E W_f + b_f) E
    \end{aligned}
\end{equation}

\noindent Here, $F_g$ denotes the forget gate, with $W_f$ and $b_f$ representing its weights and bias, respectively.

\begin{equation}
    \label{eq37}
    H^{(s)} = E' + H
\end{equation}

The model combines the filtered embeddings $E'$ with the semantic features $H$ through an additive fusion method, generating the final semantic features $H^{(s)}$. Subsequently, the information from the three modalities is concatenated to form multimodal features:

\begin{equation}
    \label{eq38}
    H^{(m)} = H^{(s)} \otimes H^{(p)} \otimes H^{(g)}
\end{equation}

\noindent Here, $\otimes$ denotes the vector concatenation operation, and $h^{(p)}_i$, $h^{(g)}_i$, $h^{(s)}_i$ represent phonetic, graphemic, and semantic feature vectors, respectively. As shown in Equation \ref{eq39}, we use a linear fusion layer to reduce the 902-dimensional vector to 768-dimensions, aligning the multimodal feature vector dimension with the original Bert dimensions:

\begin{equation}
    \label{eq39}
    \mathbb{H} = W^{(m)}\cdot H^{(m)} + b^{(m)}
\end{equation}

\noindent Here, $W^{(m)}$ and $b^{(m)}$ are the weight and bias parameters, and $\mathbb{H}$ is the fused feature vector.

\textbf{Learning Strategy:} We design a novel output pattern combining the Focal Loss function. To be specific, for the input text sequence $X=\{x_0, x_1, \cdots, x_n\}$, the corresponding labels are $Y=\{y_0, y_1, \cdots, y_n\}$. In this study, the correct values in the label set $Y$ are fixed to produce $Y'=\{y'_0, y'_1, \cdots, y'_n\}$. As shown in Equation \ref{eq312}:

\begin{equation}
    \label{eq312}
    y'_i = \begin{cases}
    1 ~~~~ \text{if} ~~ x_i = y_i \\
    y_i ~~~ \text{if} ~~ x_i \neq y_i
    \end{cases}
\end{equation}

\begin{table*}[ht]
\centering
\resizebox{0.9\textwidth}{!}{ 
\begin{tabular}{ccccccccccccc}
\hline
                                 & \multicolumn{3}{c}{\textbf{SIGHAN13}}         & \multicolumn{3}{c}{\textbf{SIGHAN14}}         & \multicolumn{3}{c}{\textbf{SIGHAN15}}         & \multicolumn{3}{c}{\textbf{CSCD-NS}}                                                                               \\ \cline{2-13} 
\multirow{-2}{*}{\textbf{Model}} & \textbf{Pre}    & \textbf{Rec}    & \textbf{F1}   & \textbf{Pre}    & \textbf{Rec}    & \textbf{F1}   & \textbf{Pre}    & \textbf{Rec}    & \textbf{F1}   & \textbf{Pre}                           & \textbf{Rec}                           & \textbf{F1}                          \\ \hline
ChineseBert                      & 82.3          & 77.1          & 79.6          & 63.3          & 66.5          & 64.9          & 69.3          & 74.1          & 71.6          & { 31.9}          & 31.3                                 & { 31.6}          \\ 
PLOME                            & /             & /             & /             & /             & /             & /             & 75.3          & 79.3          & 77.2          & 36.6                                 & 36.2                                 & 36.4                                 \\
ReaLiSe                          & \textbf{87.2} & 81.2          & 84.1          & 66.3          & 70.0          & 68.1          & 75.9          & 79.9          & 77.8          & { 36.6}          & 37.3                                 & { 36.9}          \\ 
SCOPE                            & 86.3          & 82.4          & 84.3          & \textbf{68.6}          & 71.5          & \textbf{70.2}          & \textbf{79.2} & 82.3          & 80.7          & { 43.2}          & {40.7}          & { 41.9}          \\ \hline
Deepseek-V3                      & 60            & 58.8          & 59.4          & 55.3          & 53.0          & 54.1          & 56.5          & 58.7          & 57.6          & { 54.7} & { \textbf{57.6}} & { \textbf{56.1}} \\
Chatgplm3-6B                     & 32.2          & 34.3          & 33.2          & 24.8          & 23.1          & 23.9          & 31.1          & 32.3          & 31.7          & { 34.8}          & { 33.6}          & { 34.2}          \\
Gpt-4o                           & 52.5          & 50.2          & 51.3          & 48.4          & 47.0          & 47.7          & 51.9          & 54.6          & 53.2          & { 53.5}          & { 51.2}          & { 52.3}          \\ \hline
NamBert                          & 86.4          & \textbf{82.8} & \textbf{84.5} & 66.1 & \textbf{72.5} & 69.2 & 77.5          & \textbf{84.8} & \textbf{81.0} & { \textbf{55.0}}          & { 54.2}          & { 54.6}          \\ \hline
\end{tabular}
}
\caption{Sentence-level performance on the test sets of SIGHAN and CSCD-NS, where precision (Pre), recall (Rec),
F1 (F1) for correction is reported (\%). For the SIGHAN13 dataset, the preprocessing strategy proposed by REALISE was applied. \textbf{Bold fonts} in the table indicate the best performance for that metric in the row. Baseline model results on the SIGHAN dataset are taken from their respective papers, and the LLM results are based solely on prompt strategy for the CSC task. "/" indicates that the authors have not released experimental results.}
\label{tbsighan}
\end{table*}

After model prediction, each character outputs a corresponding probability distribution $P^{x_i}=\{p^{x_i}_0, p^{x_i}_1, \cdots, p^{x_i}_m\}$, where $m$ represents the size of the dictionary, and $p^{x_i}_j$ denotes the probability of character $x_i$ being corrected to the $j$th character in the dictionary. For the input text sequence $X$ with corresponding labels $Y'$, the model outputs a probability value sequence $P=\{p_{y'_0}, p_{y'_1}, \cdots, p_{y'_n}\}$, where $p_{y'_i}$ represents the probability value of $x_i$ in the output probability distribution $P^{x_i}$. After obtaining the output probability distribution $P$, the Focal Loss function is used for loss computation. By reducing the loss weight of the correct word index "1" and increasing the weights of other indices, the model can focus on correcting errors.

\begin{equation}
    \label{eq313}
    \mathrm{FL}(P) = \sum_{i=0}^n -\alpha_{y'_i}(1-p_{y'_i})^\gamma \log(p_{y'_i})
\end{equation}

\noindent Here, $\alpha_{y'_i}$ represents the loss weight corresponding to $y'_i$.

\section{Experiments}
In this section, we present the experiments and results of NamBert on the SIGHAN dataset and the CSCD-NS dataset. Through comparative experiments with traditional multimodal and LLMs approaches, we thoroughly analyze their strengths and weaknesses in the Chinese spelling correction task. Additionally, we conduct ablation experiments to further validate the effectiveness of the proposed method.

\subsection{Experimental Results and Analysis}

As depicted in Table \ref{tbsighan}, NamBert outperforms existing multimodal models on the SIGHAN dataset. However, all models performed noticeably worse on the SIGHAN14 dataset compared to the other two datasets. Upon a detailed analysis of the model's error correction results and the dataset, we found that the SIGHAN14 dataset contains numerous annotation and sentence issues, which hindered the model's performance. This led to poor generalization and error correction capability when training and testing the CSC task with this dataset. The low quality of the SIGHAN dataset tends to mislead models, resulting in poor generalization; thus, its practical use is not ideal. Therefore, constructing a high-quality Chinese spelling correction dataset is particularly important.

When comparing the performance of LLMs and multimodal correction methods on the SIGHAN dataset, we found that although LLMs achieved good results with prompt strategies, they tended to optimize sentence expression, leading to a higher probability of over-correction. DeepSeek demonstrated significantly better performance than ChatGLM in Chinese spelling correction. The stronger the model's ability to understand Chinese information, the more accurate the generated correction results. However, LLM models generally face issues such as longer inference times, higher over-correction probability, and high fine-tuning and deployment costs.

Analyzing the results on the CSCD-NS dataset, we found that the performance of the multimodal correction model decreased due to the lack of pretraining specifically for this dataset. This highlights that traditional pre-trained language models are highly dependent on fine-tuning with correction-specific data, and improving the quality of this data is crucial for enhancing model performance. When faced with entirely new test data, LLMs showed relatively stable performance, with DeepSeek and GPT-4o gaining a slight advantage in some aspects. LLMs' strong generalization ability and stable error correction capability are distinct advantages that traditional pre-trained language models do not possess. Therefore, a key direction for improving CSC tasks is how to combine the strengths of both traditional models and LLM approaches.

%tab4
\begin{table}[ht]
\centering
\resizebox{0.9\linewidth}{!}{ 
\begin{tabular}{cccc}
\hline
\multirow{2}{*}{\textbf{Method}} & \multicolumn{3}{c}{\textbf{SIGHAN15}}               \\ \cline{2-4} 
                                 & \textbf{Pre}      & \textbf{Rec}      & \textbf{F1}     \\ \hline
w/o Multimodal                   & 76.0            & 81.7           & 78.4           \\ 
w/o Focal Loss                       & 77.1           & 83.4           & 80.1           \\ 
w/ Front Fusion                     & 75.6           & 82.9           & 79.1           \\ 
w/ Align                            & 78.4           & 81.7           & 80.0             \\ \hline
NamBert                          & \textbf{77.5} & \textbf{84.8} & \textbf{81.0} \\ \hline
\end{tabular}
}
\caption{Ablation experiment results of the NamBert
model on the SIGHAN15 test set.}
\label{tb4}
\end{table}

\subsection{Ablation Study}
Through ablation experiments, we explored the effectiveness of different modules in NamBert. The results were validated on the SIGHAN 2015 dataset, as shown in Table \ref{tb4}. The results indicate that multiple factors influence the model performance.
Firstly, multimodal information significantly contributes to the model's performance improvement. When multimodal information is removed, the F1 score drops by 2.6\%, demonstrating the crucial role of multimodal information in enhancing the model's correction accuracy. Secondly, the model's ability to effectively utilize error information is also critical. Focal Loss, by adjusting the weight of positive and negative samples, enhances the model's focus on erroneous characters, improving overall performance.

Moreover, when using the front fusion method for multimodal information fusion, the prediction layer receives less multimodal information compared to NamBert's fusion approach, resulting in a decrease in error correction performance. This indicates that NamBert's fusion strategy makes better use of multimodal information, which helps improve the model's correction ability. When the direct addition fusion method is employed, strong features override weak feature information, causing the model to lose substantial multimodal information and consequently reducing the error correction performance.

\section{Conclusion}
In this paper, we primarily explored how to enhance phonetic and graphemic information utilization in multimodal Chinese spelling correction models. Through a series of experimental analyses, we found that multimodal information is crucial for the CSC task, and effectively retaining phonetic and graphemic information is key to improving Chinese spelling correction performance. To this end, we proposed the MACU investigation experiment, which quantifies the model's ability to utilize phonetic and graphemic information through specific metrics. We introduced a new multimodal correction model, NamBert. Additionally, we conducted a comprehensive comparison with current mainstream LLMs, offering a detailed analysis of the advantages and disadvantages of different approaches. Our findings indicate that while LLMs demonstrate more stable performance, they also face challenges such as long inference times and over-correction issues. On the other hand, traditional models' high reliance on data and various modalities becomes a limiting factor in enhancing their error correction performance and generalization ability. Although LLMs have shown remarkable performance in several natural language processing tasks, solely relying on prompting strategies for CSC tasks is not yet an ideal solution. Therefore, combining the strengths of both approaches is a wise choice for further developing CSC tasks.

\section*{Limitations}
This paper primarily explores how to enhance the performance of multimodal spelling correction models. However, due to the limited number of available open-source multimodal correction models, the scope of our experimental exploration is constrained. Additionally, since high-quality Chinese spelling correction datasets are still relatively scarce, our experiments were conducted on a limited number of datasets. This may lead to the results not fully reflecting the model's performance.

\section*{Acknowledgments}
%The National Natural Science Foundation of China supports this work. 
We want to sincerely thank the reviewers for their thorough evaluation and valuable suggestions, which helped us improve the quality of this work.

\bibliography{custom}

\appendix

\section{Experimental setup}

\subsection{Datasets and Metrics}
\label{Experiments:DataMet}
This paper uses ReaLiSe's post-processing data as the training dataset, which includes SIGHAN13, SIGHAN14, SIGHAN15, and Wang271K. However, due to semantic incoherence and numerous annotation errors in the SIGHAN dataset, we additionally introduce the CSCD-NS test dataset for a more comprehensive evaluation of the model's performance. The batch size is set to 32, and the learning rate is set to 2e-4. We use widely adopted sentence-level evaluation metrics, including Precision, Recall, and F1 score.

\subsection{Baseline Models}
\label{Experiments:Baseline}
ChineseBert: This model is fine-tuned directly on the Chinese spelling correction dataset.

PLOME: Enhances correction ability by incorporating phonetic and graphemic features in the embedding layer, along with an auxiliary task of predicting phonetics.

ReaLiSe: Integrates phonetic, semantic, and graphemic features of Chinese characters using a forget gate and a three-layer Transformer encoder.

SCOPE : Uses an auxiliary phonetic prediction task to enable the semantic encoder to encode phonetic information.

LLM: For the CSC task, we conduct experiments with large language models such as ChatGLM3-6B  \footnote{\url{https://github.com/THUDM/ChatGLM3}}, GPT-4o\footnote{\url{https://openai.com/index/hello-gpt-4o}}, and DeepSeek-V3\footnote{\url{https://www.deepseek.com}}, using prompt strategy to analyze their performance in spelling correction tasks.

\end{document}